\documentclass[runningheads]{llncs}
\usepackage{graphicx}
\usepackage{tikz}
\usepackage{comment}
\usepackage{amsmath,amssymb} % define this before the line numbering.
\usepackage{color}

\usepackage[accsupp]{axessibility}  % Improves PDF readability for those with disabilities.

\graphicspath{{figures/}}

\def\ie{\emph{i.e.}\,}

\usepackage{multirow}
\usepackage{booktabs}
\usepackage{arydshln}
\newcommand{\ra}[1]{\renewcommand{\arraystretch}{#1}}
\usepackage{hyperref}
\hypersetup{
	colorlinks=true
}

\begin{document}
% \linenumbers
\pagestyle{headings}
\mainmatter
\def\ECCVSubNumber{3821}  % Insert your submission number here

\title{Multi-Granularity Prediction for Scene Text Recognition} % Replace with your title

\begin{comment}
\titlerunning{ECCV-22 submission ID \ECCVSubNumber} 
\authorrunning{ECCV-22 submission ID \ECCVSubNumber} 
\author{Anonymous ECCV submission}
\institute{Paper ID \ECCVSubNumber}
\end{comment}
%******************

% \begin{comment}
\titlerunning{Multi-Granularity Prediction for Scene Text Recognition}
\author{Peng Wang \thanks{Equal contribution. $\dagger$ Corresponding author.} \and
Cheng Da$^{\star}$ \and
Cong Yao$^{\dagger}$  }

\authorrunning{P. Wang et al.}
\institute{Alibaba DAMO Academy, Beijing, China\\
\email{\{wdp0072012,dc.dacheng08,yaocong2010\}@gmail.com}}
% \end{comment}
%******************
\maketitle

\begin{abstract}
Scene text recognition (STR) has been an active research topic in computer vision for years. To tackle this challenging problem, numerous innovative methods have been successively proposed and incorporating linguistic knowledge into STR models has recently become a prominent trend. In this work, we first draw inspiration from the recent progress in Vision Transformer (ViT) to construct a conceptually simple yet powerful vision STR model, which is built upon ViT and outperforms previous state-of-the-art models for scene text recognition, including both pure vision models and language-augmented methods. To integrate linguistic knowledge, we further propose a Multi-Granularity Prediction strategy to inject information from the language modality into the model in an \textit{implicit} way, \ie, subword representations (BPE and WordPiece) widely-used in NLP are introduced into the output space, in addition to the conventional character level representation, while no independent language model (LM) is adopted. The resultant algorithm (termed MGP-STR) is able to push the performance envelop of STR to an even higher level. Specifically, it achieves an average recognition accuracy of $93.35\%$ on standard benchmarks. Code is available at~\url{https://github.com/AlibabaResearch/AdvancedLiterateMachinery/tree/main/OCR/MGP-STR}.
%Specifically, on standard benchmarks (such as IIIT-5K, ICDAR 2015, SVT and CUTE), it achieves an average recognition accuracy of $93.35\%$. \textcolor{blue}{Code will be available soon}.
\end{abstract}

%%%%%%%%% BODY TEXT

\section{Introduction}  \label{sec:intro}
%\vspace{-1mm}
Reading text from natural scenes is one of the most indispensable abilities when building an automated machine with high-level intelligence. This explains the reason why researchers from the computer vision community sedulously have explored and investigated this complex and challenging task for decades. Scene text recognition (STR) involves decoding textual content from natural images (usually cropped sub images), which is a key component in text reading pipelines.

Previously, a number of methods~\cite{CRNN,Focusing,ASTER,MASTER} have been proposed to address the problem of scene text recognition. Recently, there emerges a new trend that linguistic knowledge is introduced into the text recognition process. SRN~\cite{SRN} devised a global semantic reasoning module (GSRM) to model global semantic context. ABINet~\cite{ABInet} proposed bidirectional cloze network (BCN) as the language model to learn bidirectional feature representation. Both SRN and ABINet adopt an independent and separate language model to capture rich language prior. 

\begin{figure}[!htp]\centering
\vspace{-2mm}
 \includegraphics[width=0.85\textwidth]{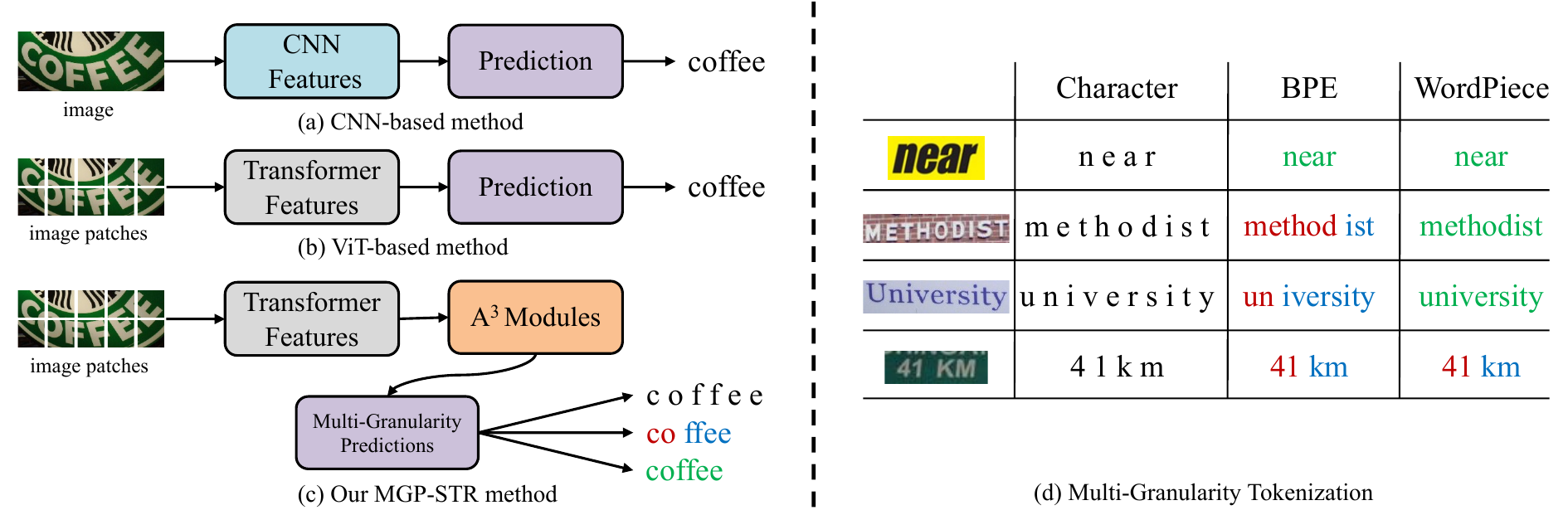}
\vspace{-2mm}
 \caption{Pipelines of classic CNN-based, ViT-based and the proposed MGP-STR scene text recognition methods are illustrated in (a), (b) and (c), respectively. (d) Examples of Character, BPE and WordPiece subword tokenization. (Best viewed in color.) }
 \label{fig:motivation}
\vspace{-2mm}
\end{figure}
In this paper, we propose to integrate linguistic knowledge in an \textbf{\textit{implicit}} way for scene text recognition. Specifically, we first construct a pure vision STR model based on ViT~\cite{dosovitskiy2020image} and a tailored Adaptive Addressing and Aggregation (A$^3 $) module inspired by TokenLearner~\cite{tokenlearner}. This model serves as a strong baseline, which already achieves better performance than previous methods for scene text recognition, according to the experimental comparisons. To further make use of linguistic knowledge to enhance the vision STR model, we explore a Multi-Granularity Prediction (MGP) strategy to inject information from the language modality. The output space of the model is expanded that subword representations (BPE and WordPiece) are introduced, \ie, the augmented model would produce two extra subword-level predictions, besides the original character-level prediction. Notably, there is no independent and separate language model. In the training phase, the resultant model (named MGP-STR) is optimized 
with a standard multi-task learning paradigm (three losses for three types of predictions) and the linguistic knowledge is naturally integrated into the ViT-based STR model. In the inference phase, the three types of predictions are fused to give the final prediction result. Experiments on standard benchmarks verify that the proposed MGP-STR algorithm can obtain state-of-the-art performance. Another advantage of MGP-STR is that it does not involve iterative refinement, which could be time-consuming in the inference phase. The pipeline of the proposed MGP-STR algorithm as well as that of previous CNN-based and ViT-based methods are shown in Fig.~\ref{fig:motivation}. In a nutshell, the major difference between MGP-STR and other methods is that it generates three types of predictions, representing textual information at different granularities: from individual characters to short character combinations, and even whole words. %while other methods only output single-granularity prediction.

The contributions of this work are summarized as follows: (1) We construct a pure vision STR model, which combines ViT with a specially designed A$^3 $ module. It already outperforms existing methods. (2) We explore an implicit way for incorporating linguistic knowledge by introducing subword representations to facilitate multi-granularity prediction, and prove that an independent language model (as used in SRN and ABINet) is not indispensable for STR models. (3) The proposed MGP-STR algorithm achieves state-of-the-art performance.

\section{Related Work}
Scene Text Recognition (STR) is a long-term subject of attention and research~\cite{zhu2016scene,long2021scene,chen2021text}. With the popularity of deep learning methods~\cite{vgg,resnet,rnn1}, its effectiveness in the field of STR has been extensively verified. Depending on whether linguistic information is applied, we roughly divide STR methods into two categories, \ie, language-free and language-augmented methods.

\subsection{Language-Free STR Methods}

The mainstream way for image feature extraction in STR methods is CNN~\cite{vgg,resnet}.
For example, previous STR methods~\cite{CRNN,rare,rnn1} utilize VGG.
%such as  CRNN~\cite{CRNN}, RARE~\cite{rare}, R$^2$AM~\cite{rnn1} and so on. 
Current STR methods~\cite{Rosetta,STAR-Net,deep,GCRNN} employ ResNet~\cite{resnet} for better performance.
Based on the powerful CNN features, various methods~\cite{zhang2022context,PIMNet,liu2022perceiving} are proposed to tackle the STR problem.
CTC-based methods~\cite{CRNN,wan20192d,STAR-Net,gtc,ctc2} use the Connectionist Temporal Classication (CTC)~\cite{ctc} to accomplish sequence recognition.
%Attention-based methods are built upon the encoder-decoder framework with RNN-attention mechanism and perform the recursive decoding process. 
Segmentation-based methods~\cite{seg,Vocabulary,TextSpotter,TextScanner} cast STR as a semantic segmentation problems. %Graph-based methods~\cite{PREN2D,he2022visual} introduce graph representation for text recognition. 

Inspired by the great success of Transformer~\cite{trans} in natural language processing (NLP) tasks, the application of Transformer in STR has also attracted more attention. Vision Transformer (ViT)~\cite{dosovitskiy2020image} that directly processes image patches without convolutions opens the beginning of using Transformer blocks instead of CNNs to solve computer vision problems~\cite{liu2021swin,SegFormer}, leading to prominent results. ViTSTR~\cite{ViTSTR} attempts to simply leverage the feature representations of the last layer of  ViT for parallel character decoding. In general, language-free methods often fail to recognize low-quality images due to the lack of language information.

\subsection{Language-Augmented STR Methods}

Obviously, language information is favourable to the recognition of low-quality images.
RNN-based methods~\cite{CRNN,rnn1,GCRNN}  can effectively capture the dependency between sequential characters, which can be regarded as an implicit language model. 
However, they cannot execute decoding in parallel during training and inference.
Recently, Transformer blocks are introduced into CNN-based framework to facilitate language content learning. 
SRN~\cite{SRN} proposes a Global Semantic Reasoning Module (GSRM) to capture the global semantic context through multiple parallel transmissions.
ABINet~\cite{ABInet} presents a Bidirectional Cloze Network (BCN) to explicitly model the language information, which is further used for iterative correction.
VisionLAN~\cite{vlan} proposes a visual reasoning module that simultaneously captures visual and language information by masking input images at the feature level. 
The mentioned above approaches utilize a specific module to integrate language information.
%While we employ the same framework with multiple classification heads to capture visual and language contents.
Meanwhile, most works~\cite{MJ1,ABInet} capture semantic information based on character-level or word-level.
In this paper, we manage to utilize multi-granularity (character, subword and even word) semantic information based on BPE and WordPiece tokenizations.

\begin{figure}[t]\centering
 \includegraphics[width=0.8\textwidth]{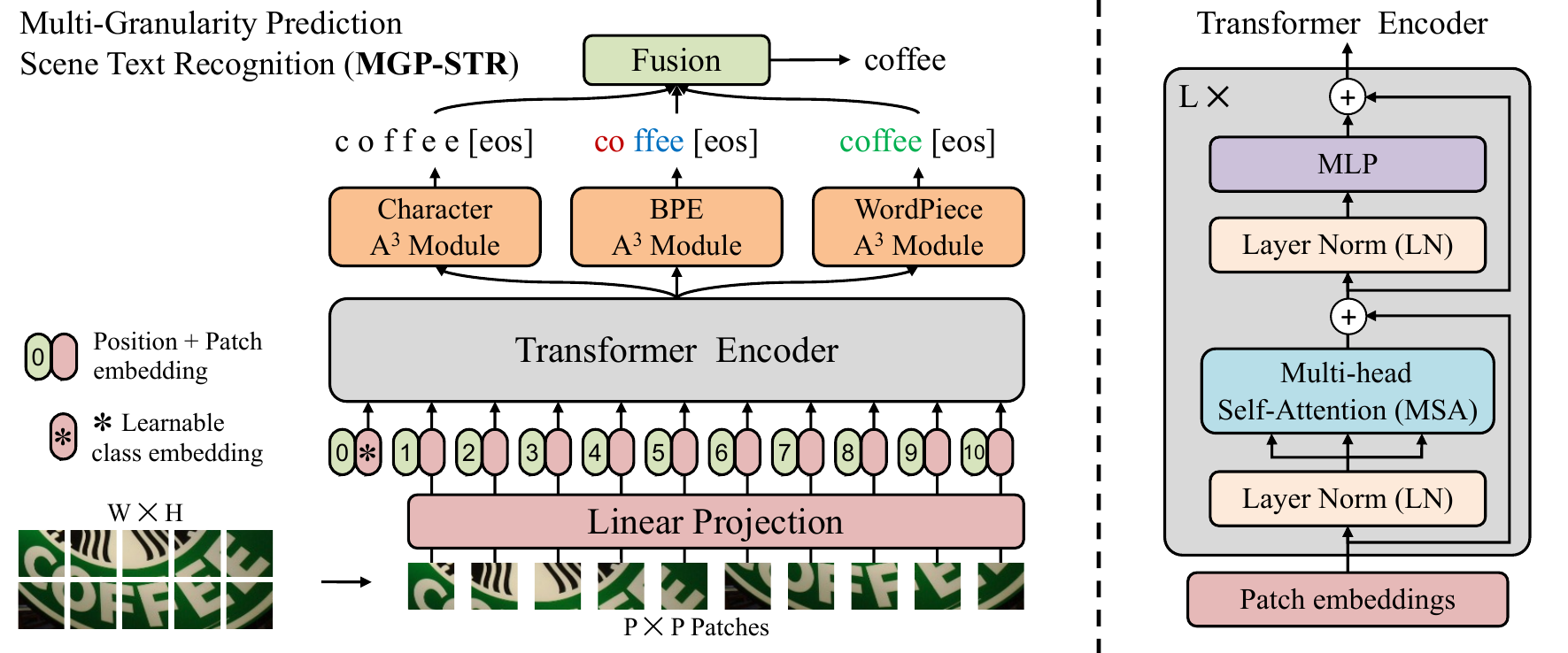}
 \caption{The architecture of the proposed MGP-STR algorithm. }
 \label{fig:overview}
\end{figure}

%-------------------------------------------------------------------------
\section{Methodology}

The overview of the proposed MGP-STR method is depicted in Fig.~\ref{fig:overview}, which is mainly built upon the original Vision Transformer (ViT) model~\cite{dosovitskiy2020image}. We propose a tailored Adaptive Addressing and Aggregation (A$^3$) module to select a meaningful combination of tokens from ViT and integrate them into one output token corresponding to a specific character, denoted as Character A$^3$ module.
Moreover, subword classification heads based on BPE A$^3$ module and WordPiece A$^3$ module are devised for subword predictions, so that the language information can be implicitly modelled. 
Finally, these multi-granularity predictions are merged via a simple and effective fusion strategy. 
\subsection{Vision Transformer Backbone}
The fundamental architecture of MGP-STR is Vision Transformer~\cite{dosovitskiy2020image,deit}, where the original image patches are directly utilized for image feature extraction by linear projection. 
As shown in Fig.~\ref{fig:overview}, an input RGB image  $ \mathbf{x} \in  \mathbb{R}^{H \times W \times C} $  is split into non-overlapping patches.
Concretely, the image is reshaped  into a sequence of flattened 2D patches $ \mathbf{x}_p \in  \mathbb{R}^{N \times (P^2 C)} $,
where $(P \times P) $ is the resolution of each image patch and $(P^2 C)$ is the number of feature channels of $ \mathbf{x}_p$. 
In this way, a 2D image is represented as a sequence with $N = HW/P^2$ tokens, which serve as the effective input sequence of Transformer blocks.
Then, these tokens of $ \mathbf{x}_p$  are linear transcribed into $D$ dimension patch embeddings. 
Similar to the original ViT~\cite{dosovitskiy2020image} backbone, a learnable  $[class]$ token embedding with $D$ dimension is introduced into patch embeddings.
And position embeddings are also added to each patch embedding to retain the positional information,
where the standard learnable 1$D$ position embedding is employed.
Thus, the generation of patch embedding vector is formulated as follows:
\begin{equation}
\begin{split}
\mathbf{z}_0=[\mathbf{x}_{class}; \mathbf{x}^1_p\mathbf{E}; \mathbf{x}^2_p\mathbf{E}; \ldots; \mathbf{x}^N_p\mathbf{E}] + \mathbf{E}_{pos},
\end{split}
\end{equation}
where $\mathbf{x}_{class} \in \mathbb{R}^{ 1 \times D}$ is the $[class]$ embedding,  $\mathbf{E}  \in \mathbb{R}^{ (P^2 C) \times D } $ is a linear projection matrix and  $ \mathbf{E}_{pos} \in \mathbb{R}^{ (N+1) \times D }   $ is the position embedding.

The resultant feature sequence $\mathbf{z}_0  \in \mathbb{R}^{ (N+1) \times D} $ serves as the input of Transformer encoder blocks,
which are mainly composed of Multi-head Self-Attention (MSA), Layer Normalization (LN), Multilayer Perceptron (MLP) and residual connection as in Fig.\ref{fig:overview}. The Transformer encoder block is formulated as:
\begin{equation}
\begin{split}
&\mathbf{z}_{l}^{\prime}=\text{MSA} (LN(\mathbf{z}_{l-1}))+ \mathbf{z}_{l-1}   \\
&\mathbf{z}_{l}=\text{MLP} (LN(\mathbf{z}_{l}^{\prime}))+ \mathbf{z}_{l}^{\prime}.
\end{split}
\end{equation}
Here, $L$ is the depth of Transformer block and  $ l=1 \ldots L $.
The MLP consists of two linear layers with GELU activation.
Finally, the output embedding $ \mathbf{z}_{L} \in \mathbb{R}^{ (N+1) \times D }$ of Transformer is utilized for subsequent text recognition.

\begin{figure}[t]\centering
 \includegraphics[width=0.8\textwidth]{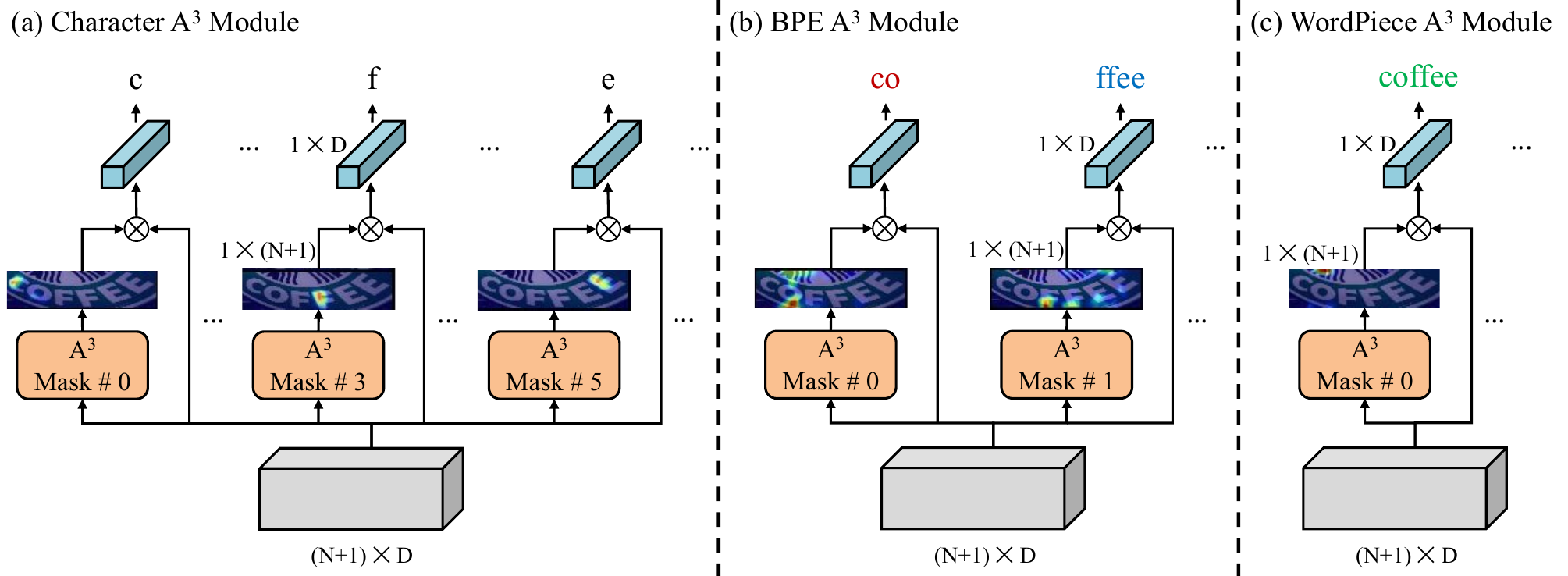}
 \caption{The detailed architectures of the three A$^3$ modules. }
 \label{fig:token}
%\vspace{-4mm}
\end{figure}

\subsection{ Adaptive Addressing and Aggregation (A$^3$) Modules} \label{sec:token}

Traditional Vision Transformers~\cite{dosovitskiy2020image,deit} usually prepend a learnable $\mathbf{x}_{class}$ token to the sequence of patch embeddings, which directly collects and aggregates the meaningful information and serves as the image representation for the classification of the whole image.
While the task of scene text recognition aims to produce a sequence of character predictions, where each character is only related to a small patch of the image.
Thus, the global image representation $ \mathbf{z}_{L}^0 \in \mathbb{R}^{ D } $  is inadequate for text recognizing task.
ViTSTR~\cite{ViTSTR} directly employs the first $T$ tokens of  $ \mathbf{z}_{L} $ for text recognition, where  $T$ is the maximum text length. Unfortunately, the rest tokens of $ \mathbf{z}_{L} $ are not fully utilized.

In order to take full advantage of the rich information of the sequence $ \mathbf{z}_{L} $ for text sequence prediction, 
we propose a tailored Adaptive Addressing and Aggregation (A$^3$)  module to select a  meaningful combination of tokens $ \mathbf{z}_{L} $ and integrate them into one token corresponding to a specific character.
Specifically, we manage to learn $T$ tokens $\mathbf{Y} = [\mathbf{y}_i]_{i=1}^{T}$ from the sequence $ \mathbf{z}_{L} $ for the subsequent text recognizing task.
An aggregation function is, thus, formulated as $\mathbf{y}_i = A_i(\mathbf{z}_{L})$, which converts the input  $ \mathbf{z}_{L} $ to a token vector $\mathbf{y}_i: \mathbb{R}^{ (N+1) \times D}  \mapsto  \mathbb{R}^{ 1 \times D } $. And  such $ T $ functions are constructed for the sequential output of text recognition.
Typically, the aggregation function $A_i(\mathbf{z}_{L})$ is implemented via a spatial attention mechanism~\cite{tokenlearner} to adaptively select the tokens from $ \mathbf{z}_{L} $ corresponding  to $i_{th}$ character.
Here, we employ function $\alpha_i(\mathbf{z}_{L})$ and softmax function to generate precise spatial attention 
mask $\mathbf{m}_i \in  \mathbb{R}^{ (N+1) \times 1}  $ from $\mathbf{z}_{L} \in \mathbb{R}^{ (N+1) \times D}$.
Thus, each output token $\mathbf{y}_i $ of A$^3$ module is produced by 
\begin{equation}
\begin{split}
&\mathbf{y}_{i}= A_i(\mathbf{z}_{L}) = \mathbf{m}_i^T \tilde{\mathbf{z}}_{L}  = \text{softmax}(\alpha_i(\mathbf{z}_{L}))^T (\mathbf{z}_{L}\mathbf{U})^T.\\
\end{split}
\end{equation}
Here, $\alpha_i$(·) is implemented by group convolution with one $1 \times 1$ kernel. And $ \mathbf{U}\in \mathbb{R}^{ D \times D}$  is a linear mapping matrix for learning feature $ \tilde{\mathbf{z}}_{L}$. Therefore, the resulting tokens of different aggregation functions are gathered together to form the final output tensor as follows:
\begin{equation}
\label{eq:Y}
\begin{split}
&\mathbf{Y}= [\mathbf{y}_{1}\mathbf{y}_{2};\ldots;\mathbf{y}_{T}] = [A_1(\mathbf{z}_{L}); A_2(\mathbf{z}_{L}); \ldots ;A_T(\mathbf{z}_{L}) ].\\
\end{split}
\end{equation}

Owing to the effective and efficient  A$^3$ module, the ultimate
representation of the text sequence is denoted as $\mathbf{Y} \in \mathbb{R}^{ T \times D}$ in Eq.~(\ref{eq:Y}). Then, a character classification head is built by $ \mathbf{G} = \mathbf{YW}^T \in \mathbb{R}^{ T \times K} $ for text sequence recognition, where $\mathbf{W} \in \mathbb{R}^{ K \times D}$ is a linear mapping matrix, $K$ is the number of categories and $ \mathbf{G} $ is the classification logist. 
We regard this module as Character  A$^3$ for character-level prediction, of which the detailed structure is illustrated in Fig.~\ref{fig:token}(a). 

\subsection{Multi-Granularity Predictions}
\label{sec:MGP}
Character tokenization that simply splits text into characters is commonly-used in scene text recognition methods.
However, this naive and standard way ignores the language information of text. 
In order to effectively resort to linguistic information for scene text recognition, we incorporate subword~\cite{subword} tokenization mechanism in NLP~\cite{BERT} into text recognition method. Subword tokenization algorithms aim to decompose rare words into meaningful subwords and remain frequently used words, so that the grammatical information of word has already been captured in the  subwords. Meanwhile, since A$^3$ module is independent of Transformer encoder backbone, we can directly add extra parallel subword A$^3$ modules for subword predictions. 
In such a way, the language information can be implicitly injected into model learning for better performance. Notably, previous methods, \ie, SRN~\cite{SRN} and ABINet~\cite{ABInet}, design an explicit transformer module for language modelling, while we cast linguistic information encoding problem as a character and subword prediction task without an explicit  language model. 

Specifically, we employ two subword tokenization algorithms Byte-Pair Encoding (BPE)~\cite{BPE} and WordPiece~\cite{wordpiece}
\footnote{Considering the potential out-of-vocabulary (OOV) issue in the inference phase, we did not directly predict whole words.} 
to produce various combinations as shown in Fig.\ref{fig:motivation}(b)(c). 
%For instance, ``coffee'' is tokenized into ``co'' and ``ffee'' by BPE, while it remains the whole word ``coffee'' via WordPiece. 
Thus, BPE A$^3$ module and WordPiece A$^3$ module are proposed for subword attention.
And two subword-level classification heads are used for subword predictions.
%Finally, these multi-granularity predictions (character and subword) are merged via a simple and effective fusion strategy.
Since subwords could be whole words (such as ``coffee'' in WordPiece), subword-level and even word-level predictions can be generated by the BPE and WordPiece classification heads. 
Along with the original character-level prediction, we denote these various outputs as multi-granularity predictions for text recognition. In this way, character-level prediction guarantees the fundamental recognition accuracy, and subword-level or word-level predictions can serve as complementary results for noised images via linguistic information. 

Technically, the architecture of BPE or WordPiece A$^3$ module is the same as Character one. They are independent of each other with different parameters.
And the numbers of categories are different for different classification heads, which depend on the vocabulary size of each tokenization method.
The cross entropy loss is employed for classification.
Additionally, the mask $\mathbf{m}_{i}$ precisely indicates the attention location of the $i_{th}$ character in Character A$^3$ module, 
while it roughly shows the $i_{th}$ subword region of the image in subword A$^3$ modules, due to the higher complexity and uncertainty of learning subword splitting.

\subsection{Fusion Strategy for Multi-Granularity Results }
\label{sec:fuse}
Multi-granularity predictions (Character, BPE and WordPiece) are generated by different A$^3$ modules and classification heads. Thus, a fusion strategy is required to merge these results. At the beginning, we attempt to fuse multi-granularity information by aggregating text features  $\mathbf{Y}$ of the output of different A$^3$ modules at feature level. However, since these features are from different granularities, the $i_{th}$  token $\mathbf{y}_i^{char}$ of character level is not aligned with the $i_{th}$ token $\mathbf{y}_i^{bpe}$ (or $\mathbf{y}_i^{wp}$) of BPE level (or WordPiece level), so that these features cannot be added for fusion. Meanwhile, even if we concatenate features by [$\mathbf{Y}_i^{char}, \mathbf{Y}_i^{bpe}, \mathbf{Y}_i^{wp}$], only one character-level head can be used for final prediction. The subword information will be greatly impaired in this way, resulting in less improvement.

Therefore, decision-level fusion strategy is employed in our method. However, perfectly fusing these predictions is a challenging problem~\cite{rrlrgu}.
We, thus, propose a compromised but efficient fusion strategy based on the prediction confidences.
Specifically, the recognition confidence of each character or subword can be obtained by the corresponding classification head.
Then, we present two fusion functions $f(\cdot)$ to produce the final recognition score based on atomic confidences:
\begin{equation}
\label{eq:mean}
f_{Mean}([c_1, c_2, \ldots, c_{eos}])= \frac{1}{n} \sum_{i=1}^{eos} c_i, \\
\end{equation}
\begin{equation}
\label{eq:prod}
f_{Cumprod}([c_1, c_2, \ldots, c_{eos}])= \prod_{i=1}^{eos}\ c_i.
\end{equation}
We only consider the confidence of valid character or subword and ending symbol $eos$, and ignore padding symbol $pad$.
``Mean'' recognition score is generated by the mean value function as in Eq.~(\ref{eq:mean}). 
And ``Cumprod'' represents the score produced by cumulative product function.
Then, three recognition scores of three classification heads for one image can be obtained by $f(\cdot)$.
We simply pick the one with the highest recognition score as the the final predicted result.

%-------------------------------------------------------------------------
\section{Experiment}
\label{Sec:Experiment}
\subsection{ Datasets}

%The comparison with different methods must be based on a unified framework, which includes the use of the same training dataset and test dataset. 
For fair comparison, we use MJSynth~\cite{MJ1,MJ2} and SynthText~\cite{ST} as training data. MJSynth contains $9M$ realistic text images and SynthText includes $7M$ synthetic text images. The test dataset consists of ``regular'' and ``irregular'' datasets. The ``regular'' dataset is mainly composed of horizontally aligned text images. IIIT 5K-Words (IIIT)~\cite{IIIT} consists of 3,000 images collected on the website. Street View Text (SVT)~\cite{SVT} contains 647 test images. ICDAR 2013 (IC13)~\cite{IC13} contains 1,095 images cropped from mall pictures, but we eventually evaluate on 857 images, discarding images that contain non-alphanumeric characters or less than three characters. The text instances in the ``irregular'' dataset are mostly curved or distorted. ICDAR 2015 (IC15)~\cite{IC15} includes 2,077 images collected from Google Eyes, but we use 1,811 images without some extremely distorted images. Street View Text-Perspective (SVTP)~\cite{SVTP} contains 639 images collected from Google Street View. CUTE80 (CUTE)~\cite{CUTE} consists of 288 curved images.

% \subsection{Implementation Details}
%   \textbf{Model Configuration} Our backbone network was introduced in 3.1. Like ViTSTR~\cite{ViTSTR}, the embedding dim in MGP is also set to 768, but the patch size is set to 4. The Vision Transformer consists of 12 stacked units, where the number of heads is 12 and the number of hidden units is 768. The size of the character-level dictionary is 38, the WordPiece-level is 30522, and the BPE-level is 50257. The maximum length of the output sequence is set to 27. The height and width of the input image are 32 x 128, and we applied the image enhancement methods provided in ViTSTR~\cite{ViTSTR} designed for STR, including various image transformations such as invert, bend, blur, noise, warp, rotate, stretch /compression, perspective, shrinkage, etc. But no matter how you change the image, it doesn't change the meaning of the text in it.

% \noindent\textbf{Model Training} We use basically the same training configuration as ViTSTR~\cite{ViTSTR}, except that the input size is resized to 32x128 and the patch size is 4. DeiT's~\cite{deit} pretrained weights are automatically downloaded before training, and we used 2 NVIDIA Tesla V100 GPUs to train our model with a batch size of 128. The Adadelta~\cite{Adadelta} optimizer was used with an initial learning rate of 1, the learning rate decay strategy is Cosine Annealing LR~\cite{cosinlr}, and the training epoch is 10 epochs.

\subsection{Implementation Details}

\subsubsection{Model Configuration.} 
MGP-STR is built upon DeiT-Base model~\cite{deit}, which is composed of $12$ stacked transformer blocks.
For each layer, the number of head is $12$ and the embedding dimension $D$ is $768$. 
More importantly, square $ 224 \times 224$ images~\cite{dosovitskiy2020image,deit,ViTSTR} are not adopted in our method.
The height $H$ and width $W$ of the input image are set to $32$ and $128$.
The patch size $P$ is set to 4 and thus $N=8 \times 32 =256$ plus one $[class]$ tokens $\mathbf{z}_L \in \mathbb{R}^{257\times768}$ can be produced.
The maximum length $T$ of the output sequence $\mathbf{Y}$ of A$^3$ module is set to $27$. 
The vocabulary size $K$ of Character classification head is set to 38, including $0-9$, $a-z$, $pad$ for padding symbol and $eos$ for ending symbol.
The vocabulary sizes of BPE and WordPiece heads are set to $50,257$ and $30,522$.

%\vspace{-4mm}

\subsubsection{Model Training.} 
The pretrained weights of DeiT-base~\cite{deit} are loaded the initial weights, except the patch embedding model, due to inconsistent patch sizes.
Common data augmentation methods~\cite{randaug} for text image are applied, such as perspective distortion, affine distortion, blur, noise and rotation.
We use $2$ NVIDIA Tesla V100 GPUs to train our model with a batch size of $100$. Adadelta~\cite{Adadelta} optimizer is employed with an initial learning rate of $1$.
The learning rate decay strategy is Cosine Annealing LR~\cite{cosinlr} and the training lasts $10$ epochs.

\subsection{Discussions on Vision Transformer and A$^3$ Modules} 

\begin{table*}[t]\centering
\setlength{\tabcolsep}{1pt}
\caption{The ablation study of the proposed vision model and the accuracy comparisons with some SOTA methods based on only vision information.}
\label{tab:char}
\begin{tabular}{|l|c|c|c|c|c|c|c|c|c|c|}
\hline
Methods & Vision & Image size (Patch) &IC13&SVT  &IIIT   & IC15 & SVTP &CUTE  &AVG \\
\hline
MASTER~\cite{MASTER} &   \multirow{3}{*}{CNN}  & - &95.3 &90.60 &95.0 &79.4 &84.5 &87.5 &89.5	 \\
SRN$_V$~\cite{SRN}  &  & - &93.2 &88.1 &92.3 &77.5 &79.4 &84.7 &86.9	 \\
ABINet$_V$~\cite{ABInet}   &  & - &94.9 &90.4 &94.6 &81.7 &84.2 &86.5 &89.8 \\
\hline
MGP-STR$_{P=16}$ &  \multirow{3}{*}{ViT}   & $ 224 \times 224 (16 \times 16) $ &95.68	&91.96	&95.13	&83.88	&85.74	&90.28	&91.07	 \\
MGP-STR$_{P=4}$ &  & $ 32 \times 128 (4 \times 4) $ &96.62	&92.27	&95.40	&84.76	&86.98	&88.54	&91.58	 \\
% MGP-STR$_{Vision}$  &  & $  32 \times 128 (4 \times 4) $ &96.62 &93.05 &96.43 &85.81 &89.61 &90.28 &92.65 \\
MGP-STR$_{Vision}$  &  & $  32 \times 128 (4 \times 4)$ &96.50 &93.20 &{96.37} &86.25 &89.46 &{90.63} &92.73 \\
\hline
\end{tabular}
\end{table*}

We analyse the influence of the patch size of Vision Transformer and the effectiveness of A$^3$ module in the proposed MGP-STR method (shown in Table~\ref{tab:char}). MGP-STR$_{P=16}$ represents the model that simply uses the first $T$ tokens of $\mathbf{z}_L$ for text recognition as in ViTSTR~\cite{ViTSTR}, where the input image is reshaped to $224 \times 224$ and the patch size is set to $16 \times 16 $. In order to retain the significant information of the original text image, $32 \times 128 $ images with $4 \times 4 $ patches are employed in MGP-STR$_{P=4}$. MGP-STR$_{P=4}$ outperfrrms MGP-STR$_{P=16}$, which indicates that the standard image size of ViT~\cite{dosovitskiy2020image,deit} is incompatible with text recognition. Thus, $32 \times 128 $ images with $4 \times 4 $ patches are used in MGP-STR.

When the Character A$^3$ module is introduced into MGP-STR, denoted as MGP-STR$_{Vision}$, the recognition performance will be further improved. MGP-STR$_{P=16}$ and MGP-STR$_{P=4}$ cannot fully learn and utilize the all tokens, while the Character A$^3$ module can adaptively aggregate features of the last layer, resulting in more sufficient learning and higher accuracy. Meanwhile, compared with SOTA text recognition methods with CNN feature extractors, the proposed MGP-STR$_{Vision}$ method achieves substantially performance improvement.%These results demonstrate that MGP-STR$_{Vision}$ is a simple yet powerful baseline for text recognition.

\begin{table*}[t]\centering
\setlength{\tabcolsep}{4pt}
\ra{1}
\caption{The accuracies of MGP-STR$_{Fuse}$ with different fusion strategies.}
\label{tab:fuse}
\begin{tabular}{|l|c|c|c|c|c|c|c|c|c|}
\hline
Method & Mode &IC13&SVT  &IIIT   & IC15 & SVTP &CUTE  &AVG \\
\hline
\multirow{3}{*}{MGP-STR$_{Fuse}$}    &  Char &96.49	&93.66	&96.1	&86.14	&88.83	&89.58	&92.53	 \\
& Mean &97.31	&94.28	&96.60	&86.97	&90.23	&90.97	&93.28	 \\
& Cumprod &97.32	&94.74	&96.40	&87.24	&91.01	&90.28	&93.35	 \\
\hline
\end{tabular}
\end{table*}

\subsection{Discussions on Multi-Granularity Predictions }

\begin{table*}[t]\centering
\setlength{\tabcolsep}{1.5pt}
\ra{1}
\caption{The accuracy results of the four variants of MGP-STR model. ``Char'', ``BPE'' and ``WP'' at ``Output'' represent predictions of Character, BPE and WordPiece classification head  in each model, respectively. ``Fuse'' represents the fused results.}
\label{tab:CBW}
\begin{tabular}{|l|c|c|c|c|c|c|c|c|c|c|c|c|}
\hline
Methods & Char & BPE & WP &Output &IC13&SVT  &IIIT   & IC15 & SVTP &CUTE  &AVG \\
\hline
% MGP-STR$_{Vision}$ &\checkmark & $ \times $ & $ \times $ &  Char &96.50	&93.51	&96.13	&86.14	&89.30	&91.32	&92.65	 \\
MGP-STR$_{Vision}$ &\checkmark & $ \times $ & $ \times $ &  Char &96.50 &93.20 &{96.37} &86.25 &89.46 &{90.63} &92.73 \\
\hline
\multirow{3}{*}{MGP-STR$_{C+B}$} &  \multirow{3}{*}{\checkmark} & \multirow{3}{*}{ \checkmark} &  \multirow{3}{*}{$ \times $} &  Char &97.43	&93.82	&96.53	&85.92	&89.15	&90.28	&92.84	 \\
\cline{6-12}
& & & &  BPE &97.78	&94.13	&90.00	&81.12	&88.37	&82.64	&88.63	 \\
%\cline{6-12}
& & & &  Fuse &97.67 &94.47	&96.73	&86.97	&88.99	&89.93	&93.24	 \\
\hline
\multirow{3}{*}{MGP-STR$_{C+W}$} &  \multirow{3}{*}{\checkmark} &  \multirow{3}{*}{$ \times $} & \multirow{3}{*}{ \checkmark}  &  Char &96.97	&93.97	&96.30	&86.20	&90.39	&89.93	&92.87	 \\
%\cline{6-12}
& & & &  WP &95.92	&93.35	&87.70	&78.74	&89.30	&80.21	&86.78	 \\
%\cline{6-12}
& & & &  Fuse &97.32	&93.82	&96.60	&86.91	&90.54	&90.97	&93.25	 \\
\hline
\multirow{4}{*}{MGP-STR$_{Fuse}$} &  \multirow{4}{*}{\checkmark} &  \multirow{4}{*}{\checkmark} & \multirow{4}{*}{ \checkmark}  &  Char &96.49	&93.66	&96.10	&86.14	&88.83	&89.58	&92.53	 \\
%\cline{6-12}
& & & &  BPE &95.56	&93.66	&88.73	&79.84	&89.76	&83.33	&87.63	 \\
%\cline{6-12}
& & & &  WP &95.79	&94.59	&86.37	&77.36	&89.61	&79.86	&85.99	 \\
%\cline{6-12}
& & & &  Fuse &97.32	&94.74	&96.40	&87.24	&91.01	&90.28	&93.35	 \\
\hline
\end{tabular}
\end{table*}

\subsubsection{Effect of Fusion Strategy. }
Since the subwords generated by subword tokenization methods contain grammatical information, we directly employ subwords as the targets of our method to capture the language information implicitly. As described in Sec.~\ref{sec:token}, two different subword tokenizations (BPE and WordPiece) are employed for complementary multi-granularity predictions. 
Besides the character prediction, we propose two fusion strategies to further merge these three results, denoted as ``Mean’’ and ``Cumprod’’ as mentioned in Sec.~\ref{sec:fuse}. We denote this method that merges three results as MGP-STR$_{Fuse}$, and the accuracy results of MGP-STR$_{Fuse}$ with different fusion strategies are listed in Table~\ref{tab:fuse}. Additionally, the first line ``Char'' in Table~\ref{tab:fuse} records the result of character classification head in MGP-STR$_{Fuse}$. It is clear to see that both ``Mean’’ and ``Cumprod’’ fusion strategies can significantly improve the recognition accuracy over single character-level result. Due the better performance of ``Cumprod’’ strategy, we employ it as the fusion strategy in the following experiments.

\subsubsection{Effect of Subword Representations.}
%To further explore the effect of different subword tokenization, 
We evaluate four variants of the MGP-STR model, and the performances of these four methods are elaborately reported in Table~\ref{tab:CBW}, including the fused results and the results of each single classification. Specifically, MGP-STR$_{Vision}$ with only Character A$^3$ module has already obtained promising results. MGP-STR$_{C+B}$ and MGP-STR$_{C+W}$ incorporate Character A$^3$ module with BPE A$^3$ module and WordPiece A$^3$ module, respectively. No matter which subword tokenization is used alone, the accuracy of ``Fuse’’ can exceed that of ``Char’’ in both MGP-STR$_{C+B}$ and MGP-STR$_{C+W}$ methods, respectively. Notably, the performance of the classification  of ``BPE’’ or ``WP’’ could be better than that of ``Char’’ on SVP and SVTP datasets in the same model. These results show that subword predictions can 
boost text recognition performance by implicitly introducing language information. Thus, MGP-STR$_{Fuse}$ with three A$^3$ modules can produce complementary multi-granularity predictions (character, subword and even word). By fusing these multi-granularity results, MGP-STR$_{Fuse}$ obtains the best performance.

\subsubsection{Comparison with BCN.}
Bidirectional cloze network (BCN) is designed in ABINet~\cite{ABInet} for explicit language modelling, and it leads to favorable improvement over pure vision model. We equip MGP-STR$_{Vision}$ with BCN as a competitor of MGP-STR$_{Fuse}$ to verify the advantage of multi-granularity predictions. Concretely, we first reduce the dimension $768$ of representation feature $\mathbf{Y}$ to $512$ for feature fusion of the output of BCN. Following the training setting in~\cite{ABInet}, the model results are reported in Table~\ref{tab:BCN}.  The accuracy of ``V+L'' is  further imporved over the pure vision prediction ``V'' in MGP-STR$_{Vision}$+BCN, and better than the original ABINet~\cite{ABInet}. However, the performance of MGP-STR$_{Vision}$+BCN is a little worse than that of MGP-STR$_{Fuse}$. In addition, we provide the upper bound on the performance of MGP-STR$_{Fuse}$, denoted as MGP-STR$_{Fuse}^*$ in Table 4. If one of the three predictions (``Char'', ``BPE'' and ``WP'') is right, the final prediction is considered correct. 
The highest score of MGP-STR$_{Fuse}^*$ demonstrates the good potential of  multi-granularity predictions.
Moreover, MGP-STR$_{Fuse}$ only requires two new subword prediction heads, rather than the design of a specific and explicit language model in~\cite{ABInet,SRN}. 
%These experiments demonstrate the effectiveness of multi-granularity prediction in language modelling of scene text recognition.

\begin{table*}[t]\centering
\setlength{\tabcolsep}{4pt}
\ra{1}
\caption{The accuracy results of MGP-STR$_{Vision}$ equipped with BCN and multi-granularity prediction. ``V'' represents the results of the pure vision output. ``V+L'' represents the results based on the both vision and language parts.}
\label{tab:BCN}
\begin{tabular}{|l|c|c|c|c|c|c|c|c|c|}
\hline
Methods & Mode &IC13&SVT  &IIIT   & IC15 & SVTP &CUTE  &AVG \\
\hline
\multirow{2}{*}{MGP-STR$_{Vision}$+BCN}  &  V &96.97	&93.82	&95.90	&85.53	&89.15	&89.58	&92.40	 \\
& V+L &97.32	&95.36	&95.97	&86.69	&91.78	&89.93	&93.14	 \\
\hline
MGP-STR$_{Fuse}$   & V+L &97.32	&94.74	&96.40	&87.24	&91.01	&90.28	&93.35	 \\
MGP-STR$_{Fuse}^*$   & V+L &97.66	&96.29	&96.97	&89.06	&92.09	&92.01	&94.38	 \\
\hline
\end{tabular}
\end{table*}

\subsection{Results with Different ViT Backbones} 

\begin{table*}[t]\centering
\setlength{\tabcolsep}{5pt}
\ra{1}
\caption{The accuracy results of MGP-STR$_{Fuse}$ with different ViT backbones.}
\label{tab:y0}
\begin{tabular}{|l|c|c|c|c|c|c|c|c|c|}
\hline
Backbone & Output &IC13&SVT  &IIIT   & IC15 & SVTP &CUTE  &AVG \\
\hline
\multirow{4}{*}{DeiT-Tiny}  &  Char &93.47	&90.57	&93.93	&82.94	&81.71	&84.38	&89.36	 \\
&  BPE &87.40	&84.39	&83.17	&73.72	&77.83	&71.53	&80.48	 \\
&  WP &53.79	&45.44	&60.07	&52.57	&42.79	&42.71	&53.92	 \\
&  Fuse &94.05	&91.19	&94.30	&83.38	&83.57	&84.38	&89.91	 \\
\hline
\multirow{4}{*}{DeiT-Small}  &  Char &95.92	&91.04	&94.97	&84.59	&85.89	&86.81	&91.01	 \\
&  BPE &96.27	&93.35	&89.37	&79.74	&86.67	&82.29	&87.61	 \\
&  WP &75.50	&70.48	&74.70	&66.81	&68.06	&62.15	&71.36	 \\
&  Fuse &96.38	&93.51	&95.30	&86.09	&87.29	&87.85	&91.96	 \\
\hline
\multirow{4}{*}{DeiT-Base}  &  Char &96.49	&93.66	&96.10	&86.14	&88.83	&89.58	&92.53	 \\
&  BPE &95.56	&93.66	&88.73	&79.84	&89.76	&83.33	&87.63	 \\
&  WP  &95.79	&94.59	&86.37	&77.36	&89.61	&79.86	&85.99	 \\
&  Fuse &97.32	&94.74	&96.40	&87.24	&91.01	&90.28	&93.35	 \\
\hline
\end{tabular}
\end{table*}

All of the proposed MGP-STR models mentioned earlier are based on DeiT-Base~\cite{deit}. We also introduce two smaller models, namely DeiT-Small and DeiT-Tiny as presented in~\cite{deit} to further evaluate the effectiveness of MGP-STR$_{Fuse}$ method. Specifically, the embedding dimensions of DeiT-Small and DeiT-Tiny are reduced to $384$ and $192$, respectively. Table 5 records the results of each prediction head of the MGP-STR$_{Fuse}$ method with different ViT backbones. Clearly, fusing multi-granularity predictions can still improve the performance of pure character-level prediction in every backbone. And bigger models achieve better performance in the same  head. More importantly, the results of ``Char’’ in DeiT-Small and even DeiT-Tiny have already surpassed the SOTA pure CNN-based vision models, referring to Table~\ref{tab:char}. Therefore, MGP-STR$_{Vision}$ with small or tiny ViT backbone is also a competitive vision model and multi-granularity prediction can also work well in different ViT backbones. 

\begin{table*}[t]
\setlength{\tabcolsep}{5pt}
\ra{1}
\centering
\caption{The comparisons with SOTA methods on several public benchmarks.}
\label{tab:sota}
\begin{tabular}{|l|c|c|c|c|c|c|c|}
\hline
\multirow{2}{*}{Methods}   &\multicolumn{3}{c|}{Regular Text} &\multicolumn{3}{c|}{Irregular Text} &\multirow{2}{*}{AVG}\\
\cline{2-7}
&IC13 &SVT  &IIIT   & IC15 & SVTP &CUTE &  \\
\hline
TBRA~\cite{deep}   &93.6 &87.5 &87.9 &77.6 &79.2  & 74.0  & 84.6 \\
ViTSTR~\cite{ViTSTR} &93.2 & 87.7 & 88.4 &78.5 &81.8 &81.3 &85.6 \\
ESIR~\cite{ESIR}   &91.3  &90.2 &93.3 &76.9 &79.6 &83.3  &  87.1  \\
DAN~\cite{DAN} &93.9  &89.2 &94.3  &74.5  &80.0 &84.4   &  87.2  \\
%SAM~\cite{TextSpotter}     &95.3 &90.6 &93.9 &77.3 &82.2 &87.8  &  88.34  \\
SE-ASTER~\cite{SEED}   &92.8 &89.6 &93.8 &80.0 &81.4 &83.6  & 88.3  \\
RobustScanner~\cite{RobustScanner}     & 94.8 & 88.1  & 95.3 & 77.1 & 79.5 &  90.3 & 88.4 \\
TextScanner~\cite{TextScanner} &92.9 &90.1 &93.9 &79.4 &84.3 &83.3  & 88.5 \\
%ScRN~\cite{ScRN}  &93.9  & 88.9   &94.4    &78.7 & 80.8   & 87.5 & 88.44 \\
SATRN~\cite{SATRN}    &94.1 &91.3 &92.8  &79.0 &86.5 &87.8 &  88.6 \\
MASTER~\cite{MASTER} &95.3	&90.6	&95.0	&79.4	&84.5	&87.5 & 89.5 \\
\hline
SRN~\cite{SRN}     & 95.5 & 91.5  & 94.8  &82.7  &85.1   &87.8 & 90.4 \\
VisionLAN~\cite{vlan} &95.7 &91.7& 95.8 &83.7 &86.0 &88.5 &91.2 \\
ABINet~\cite{ABInet}    &\textbf{97.4} &{93.5} &{96.2} &{86.0} &{89.3} &89.2  & {92.6} \\
% ABINet~\cite{ABInet}    &{97.3} &94.0 &{96.4} &{85.5} &{89.1} &89.2  & {92.6} \\
\hline
% MGP-STR$_{Vision}$  &96.62 &93.05 &\textbf{96.43} &85.81 &89.61 &\textbf{90.28} &92.65 \\
MGP-STR$_{Vision}$  &96.50 &93.20 &{96.37} &86.25 &89.46 &\textbf{90.63} &92.73 \\
MGP-STR$_{Fuse}$  &{97.32}	&\textbf{94.74}	&\textbf{96.40}	&\textbf{87.24}	&\textbf{91.01}	&{90.28}	&\textbf{93.35} \\
\hline
\end{tabular}
\end{table*}

\subsection{Comparisons with State-of-the-Arts} 
\label{Sec:sota}
We compare the proposed MGP-STR$_{Vision}$ and MGP-STR$_{Fuse}$ methods with previous state-of-the-art scene text recognition methods, and the results on $6$ standard benchmarks  are summarized in Table~\ref{tab:sota}. All of compared methods and ours are trained on synthetic datasets MJ and ST for fair evaluation. And the results are obtained without any lexicon based post-processing. Generally, language-aware methods (\ie, SRN~\cite{SRN}, VisionLAN~\cite{vlan}, ABINet~\cite{ABInet} and MGP-STR$_{Fuse}$) perform better than other language-free methods, showing the significance of linguistic information. Notably, MGP-STR$_{Vision}$ without any language information has already outperformed the state-of-the-art method ABINet with explicit language model. Owing to the multi-granularity prediction, MGP-STR$_{Fuse}$ obtains more impressive results further, which outperforms ABINet with $0.7\%$ improvement on average accuracy. 

\subsection{Details of Multi-Granularity Predictions}  \label{Sec:fa}

\begin{table*}[t]\centering
\setlength{\tabcolsep}{2pt}
\ra{1}
\caption{The details of multi-granularity prediction of MGP-STR$_{Fuse}$, including the scores of each prediction head, the intermediate multi-granularity (Gra.) results and the final prediction (Pred.). Best viewed in color.   }
\label{tab:fa}
\begin{tabular}{|c|c|c|c|c|c|c|c|}
\hline
Images & GT &Output &Char  &BPE   & WP  & Fuse \\
\hline
    \multirow{3}{*}{
    \begin{minipage}[b]{0.15\columnwidth}
		\centering
		\raisebox{-.5\height}{\includegraphics[width=\linewidth]{CUTE80_tabbe_0.1643_table_0.9521_table_0.9813_table_0.9813_table.png}}
	\end{minipage}
	}
	& \multirow{3}{*}{table}
	&Score & 0.1643 & 0.9813 & 0.9521 & 0.9813\\
	\cline{3-7}
	& &Gra. & tabbe & \textcolor{green}{table} & \textcolor{green}{table}  & -\\
	\cline{3-7}
	& &Pred. & tabbe & table &  table & table \\
\hline
    \multirow{3}{*}{
    \begin{minipage}[b]{0.15\columnwidth}
		\centering
		\raisebox{-.5\height}{\includegraphics[width=\linewidth]{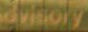}}
	\end{minipage}
	}
	& \multirow{3}{*}{dvisory}
	&Score &0.0316 & 0.8218 & 0.2574 & 0.8218\\
	\cline{3-7}
	& &Gra. &divsoory &  \textcolor{red}{d}  \textcolor{blue}{visory} &  \textcolor{green}{dvisory} & - \\
	\cline{3-7}
	& &Pred. &divsoory & dvisory  & dvisory & dvisory \\
	\hline
    \multirow{3}{*}{
    \begin{minipage}[b]{0.15\columnwidth}
		\centering
		\raisebox{-.5\height}{\includegraphics[width=\linewidth]{SVT_watercourss_0.1565_waterco_0.632_water_course_0.8295_watercourse_0.8295_watercourse.png}}
	\end{minipage}
	}
	& \multirow{3}{*}{watercourse}
	&Score &0.1565 & 0.8295 & 0.632 & 0.8295\\
	\cline{3-7}
	& &Gra. & watercourss &  \textcolor{red}{water}  \textcolor{blue}{course} &  \textcolor{green}{waterco} & - \\
	\cline{3-7}
	& &Pred. &watercourss & watercourse  & waterco & watercourse \\
		\hline
    \multirow{3}{*}{
    \begin{minipage}[b]{0.15\columnwidth}
		\centering
		\raisebox{-.5\height}{\includegraphics[width=\linewidth]{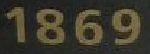}}
	\end{minipage}
	}
	& \multirow{3}{*}{1869}
	&Score &0.9999 & 0.9207 & 0.0354 & 0.9999\\
	\cline{3-7}
	& &Gra. & 1869 &  \textcolor{red}{18}  \textcolor{blue}{69} &  \textcolor{green}{18} & - \\
	\cline{3-7}
	& &Pred. &1869 & 1869  & 18 & 1869 \\
	\hline
    \multirow{3}{*}{
    \begin{minipage}[b]{0.12\columnwidth}
		\centering
		\raisebox{-.5\height}{\includegraphics[width=\linewidth]{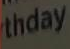}}
	\end{minipage}
	}
	& \multirow{3}{*}{thday}
	&Score &0.9998 & 0.5983 & 0.7638 & 0.9998 \\
	\cline{3-7}
	& &Gra. & thday &  \textcolor{red}{th}  \textcolor{blue}{day} &  \textcolor{green}{today} & - \\
	\cline{3-7}
	& &Pred. &thday & thday  & today & thday \\
				\hline
    \multirow{3}{*}{
    \begin{minipage}[b]{0.12\columnwidth}
		\centering
		\raisebox{-.5\height}{\includegraphics[width=\linewidth]{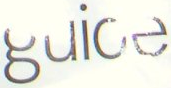}}
	\end{minipage}
	}
	& \multirow{3}{*}{guide}
	&Score &0.9675 & 0.6959 & 0.1131 & 0.9675 \\
	\cline{3-7}
	& &Gra. & guice &  \textcolor{red}{gu}  \textcolor{blue}{ice} &  \textcolor{green}{guide} & - \\
	\cline{3-7}
	& &Pred. &guice & guice  & guide & guice \\
	\hline
\end{tabular}
\end{table*}

We show the detailed prediction process of the proposed MGP-STR$_{Fuse}$ method on $6$ test images from standard datasets. In the first three images, the results of character-level prediction are incorrect, due to irregular font, motion blur and curved shape, respectively. The scores of character prediction are very low, since the images are difficult to recognize and one character is wrong in each image. However, ``BPE'' and ``WP'' heads can recognize ``table'' image with high scores. And ``BPE'' can make correct predictions with two subwords on ``dvisory'' and ``watercourse'' images, while ``WP''  is wrong in ``watercourse'' image. After fusion, the mistakes can be corrected. From the rest three images, interesting phenomena can be observed. The predictions of ``Char'' and ``BPE''  conform to the images. The predictions of ``WP'', however, attempt to produce strings with more linguistic content, like ``today'' and ``guide''.
Generally, ``Char'' aims to produce characters one by one, while ``BPE'' usually generates n-gram segments related to image and ``WP'' tends to directly predict words that are linguistically meaningful. These prove the predictions of different granularities convey text information in different aspects and are indeed complementary.

\subsection{Visualization of Spatial Attention Maps of A$^3$ Modules} 

\begin{figure}[t]\centering
 \includegraphics[width=0.8\textwidth]{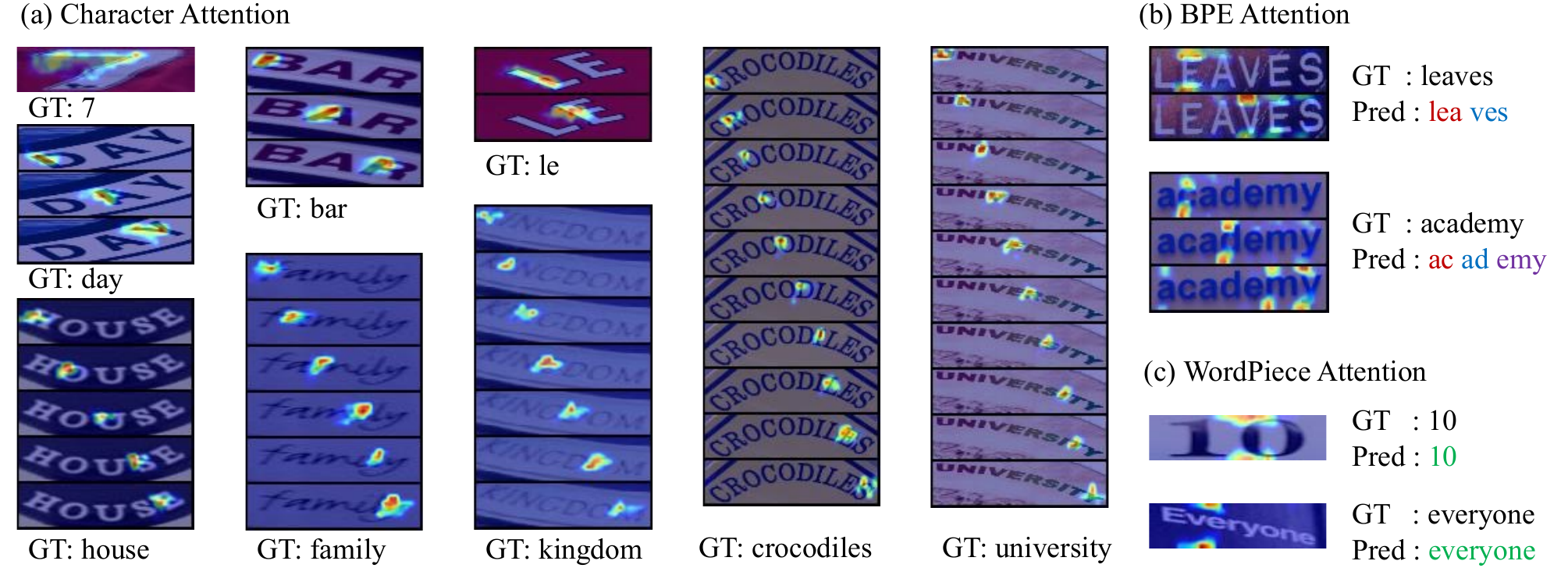}
 \caption{The illustration of spatial attention masks on Character A$^3$  module, BPE A$^3$  module and WordPiece A$^3$  module, respectively. }
 \label{fig:viscase}
%\vspace{-4mm}
\end{figure}

Exemplar attention maps $ \mathbf{m}_i $ of Character, BPE and WordPiece A$^3$ modules are shown in Fig.~\ref{fig:viscase}.
Character A$^3$ module shows extremely precise addressing ability on a variety of text images.
Specifically, for the ``7'' image with one character, the attention mask seems like the ``7'' shape.
For the ``day'' and ``bar'' images with three characters, the attention masks of middle character ``a'' are  completely different, verifying the adaptiveness of A$^3$ module.
%In short, the character attention maps can perform well on curved, hand-writing, short and even long curved images.
As depicted in Fig.\ref{fig:motivation}(d) and in Table~\ref{tab:fa}, BPE tends to generate short segments, thus the attention masks of BPE are spilt into $2$ or $3$ areas as shown in ``leaves'' and ``academy'' images.
This is probably because that performing subword splitting and character addressing simultaneously is difficult.
Moreover, WordPiece often produces a whole word, and the attention maps should be the whole feature map. Since the attention maps produced by the softmax function are usually sparse, the attention maps of WordPiece are not as appealing as those of Character A$^3$ module.
These results are consistent to those of Table~\ref{tab:CBW}, where the accuracies of ``BPE'' and ``WP'' are relatively lower than ``Char'', due to the difficulty of precise subword prediction. 
% Therefore, MGP$_{Vision}$ provides a strong fundamental baseline,
% and language information within subword are employed for better performance.

%\subsection{\textcolor{blue}{Comparison on Inference Time and Model Size } } 
\subsection{Comparisons of Inference Time and Model Size} 

\begin{table}[h]\centering
%\vspace{-2mm}
\setlength{\tabcolsep}{5pt}
\caption{Comparisons on inference time and model size.}
\label{tab:speed}
%\vspace{-2mm}
\begin{tabular}{|l|l|l|}
\hline
%\textbf{Model}  & \begin{tabular}[c]{@{}l@{}}\textbf{Speed}\\(msec/image)\end{tabular} & \begin{tabular}[c]{@{}l@{}}\textbf{Parameters}\\(1 x 10$^6$)\end{tabular}  \\
{Methods}  & {Time} (ms/image) & Parameters ($1 \times 10^6$)  \\
\hline
ABINet-S-iter1/iter2/iter3 & 13.7/18.6/24.3 & 32.8 \\
\hline
ABINet-L-iter1/iter2/iter3 & 16.1/21.4/26.8 & 36.7 \\
\hline
MGP-STR$_{Vision}$-tiny/small/base  & 10.6/10.8/10.9  & 5.4/21.4/85.5 \\
\hline
MGP-STR$_{Fuse}$-tiny/small/base  & 12.0/12.2/12.3  & 21.0/52.6/148.0 \\
\hline
\end{tabular}
%\vspace{-3mm}
\end{table}

%\textcolor{blue}{
The model sizes and latencies of the proposed MGP-STR with different settings as well as those of ABINet are depicted in Table.~\ref{tab:speed}~\footnote{All the evaluations are conducted on a NVIDIA V100 GPU.}. Since MGP-STR is equipped with a regular Vision Transformer (ViT) and involves no iterative refinement, the inference speed of MGP-STR is very fast: $12.3$ms with ViT-Base backbone. Compared with ABINet, MGP-STR runs much faster ($12.3$ms vs. $26.8$ms), while obtaining higher performance. The model size of MGP-STR is relatively large. However, a large portion of the model parameter is from the BPE and WordPiece branches. For the scenarios that are sensitive to model size or with limited memory space, MGP-STR$_{Vision}$ is an excellent choice.%The performance of MGP-STR$_{Vision}$ is higher than that of previous SOTA methods, while its model size is significantly reduced (since there are no BPE and WordPiece branches) and the running speed is further boosted. 
%}

\section{Conclusion}

We presented a ViT-based pure vision model for STR, which shows its superiority in recognition accuracy. To further promote recognition accuracy of this baseline model, we proposed a Multi-Granularity Prediction strategy to take advantage of linguistic knowledge. The resultant model achieves state-of-the-art performance on widely-used dadatsets. In the future, we will extend the idea of multi-granularity prediction to broader domains.

\bibliographystyle{splncs04}

\end{document}